\documentclass{elsart}

\usepackage{amsmath,amsfonts,amssymb}
\usepackage{latexsym}

\begin{document}

\begin{frontmatter}
  \title{Self-Organising Networks for Classification: developing Applications to
    Science Analysis for Astroparticle Physics} 
  \author{A.~De Angelis,
    P.~Boinee,~M. Frailis, E.~Milotti}   
  \address{Dipartimento di Fisica
    dell'Universit\`a di~Udine and INFN Trieste, via delle Scienze~208,
    I-33100~Udine, Italy}

\begin{abstract}
  Physics analysis in astroparticle experiments requires the capability of
  recognizing new phenomena; in order to establish what is new, it is important
  to develop tools for automatic classification, able to compare the final
  result with data from different detectors.  A typical example is the problem
  of Gamma Ray Burst detection, classification, and possible association to
  known sources: for this task physicists will need in the next years tools to
  associate data from optical databases, from satellite experiments (EGRET,
  GLAST), and from Cherenkov telescopes (MAGIC, HESS, CANGAROO, VERITAS).
\end{abstract}




\begin{keyword}
SOM \sep Classification \sep Clustering \sep GRB

  \PACS
\end{keyword}
\end{frontmatter}

\section{Introduction}
Clustering of features is an important problem in many physics experiments. Such
an analysis task can be performed:
\begin{itemize}
\item in a supervised way, when the analyst has some examples, for which the
  correct classification is known. This can be done, for example, in most
  problems related to particle physics at accelerators, where there is a
  generally good knowledge of detectors and of the underlying physics, and good
  simulations are available.
\item in an unsupervised way, when the events are partitioned into classes of
  similar elements, without using additional information. This is the case
  especially for fields operating in a discovery regime, as, e.g., astroparticle
  physics.
\end{itemize}

The idea of automatic classification is not new in particle and astroparticle
physics.  Cleaning up the signal and separating concurrent signals when
nonlinear effects and high-order correlations are important is a standard in
particle physics since the analysis of the branching fraction of the Z boson
into $b\bar{b}$ pairs by DELPHI \cite{del92}.

An important literature exists for the use of automatic classifiers in
astroparticle physics (see for example \cite{praveen} and references therein).
Such a classification was mostly done with the use of Multilayer Perceptrons,
while the bulk of the works based on unsupervised classification uses
Independent Component Analysis (see for example \cite{ciccia} and references
therein).  Studies based on Self-Organized Maps and Growing Self-Organizing
Networks \cite{p2,p3,p6,p7,p8} have recently started \cite{p16}, but a general
framework for multiwavelength classification is still missing.

\section{A case study in astroparticle physics}
Gamma-ray astroparticle physics is a relatively new science; it has as a
counterpart optical astrophysics, one of the oldest sciences.  Many of the
objects we observe in the gamma sky, sensitive to the phenomena of high-energy
physics, have an optical counterpart or clear relations to optical objects.
Finding what is a signature of a new phenomenon requires the ability to classify
observations, and the ability to recognize what is not new.

Astrophysical databases contain large amounts of data; one example is given by
the growing number of experiments studying Gamma Ray Bursts (GRBs). Data sets
can be found in several archives (see e.g.~Ref. \cite{p1}).

Large datasets are available from systematic sky surveys. The size of such
databases is now of the order of $10^{12}$ bytes, but in the near future it will
grow by three orders of magnitude thanks to the technological development of
telescopes and detectors.  Surveys are done on a wide energy range (from
$10^{-7}$ to $10^{14}$ eV), and they are heterogeneous (mission-oriented,
platform and instrument dependent). The attributes registered are variable
(polarization etc.); numerical simulations have to be matched to real data.

Such a complexity poses nontrivial data management issues (see \cite{frailis});
moreover, we need uniform interfaces to access complex data.  A few projects
started in the last years with the simple purpose of making the data readable.

\section{The project at the University of Udine}

At the University of Udine we are developing a project involving data
organization, data mining and analysis tools for the analysis of gamma sources
(Gamma Ray Bursts in particular: most of the EGRET sources were unidentified).

The sources detected by GLAST \cite{glast} and MAGIC \cite{magic} will be
compared with existing databases to detect what is new. What is new can be then
classified based on an unsupervised classifier.

Another important analysis tool is a powerful visualization package: the idea is
to visually present many variables together offering a degree of control over a
number of different visual properties.  High dimensionality of data set and
visual properties such as color, size can be added to the position property for
proper visualization purposes.  Multiple views can be used by linking all
separate views together when the use of these properties makes it difficult.

\subsection{Classification of GRBs}

The kernel of the analysis is the strategy for the classification. With the
growing number of experiments dedicated to GRBs \cite{grb} it is essential to
optimize the techniques for the complex task of classification. Artificial
Intelligence- (AI-) based pattern recognition algorithms are one possible
candidate: automated linear classification of vector data into a given number
(or an arbitrary number) of classes is a well established technique in the field
of machine learning.  Several varieties of AI-based classifiers
exist~\cite{p16}.

Clustering is the unsupervised classification of patterns~\cite{p4}
(observations, data items or feature vectors) into groups called clusters.
Clustering is useful in several exploratory pattern analysis, grouping, decision
making and machine learning situations including data mining, document
retrieval, image segmentation and pattern classification.

Self-Organising Neural Networks~\cite{p2,p3,p6,p7,p8} are often used to cluster
input data. Similar patterns are grouped by the network and are represented by a
single unit. This grouping is done automatically on the basis of data
correlations. Well-known examples of Self-Organising Artificial Neural Networks
(ANN) used for clustering include Kohonen's self-organising maps,
Self-Organising Tree Algorithm (SOTA), Growing Cell Structures (GCS).

In our prototype, Self-Organizing Maps (SOM) were used.

\section {Research Perspectives}

One promising area where the potential of self-organizing networks has not been
fully exploited is certainly data mining and knowledge discovery. Clustering
huge data sets without knowing in advance the number of clusters is something
such strategy should excel at.
 
Making hybrid neural networks (combining various self -organizing networks) can
result in an efficient clustering.

Visualisation has an important role in cluster analysis . Advanced Visualisation
techniques~\cite{p15} such as Galaxies, Correlation Tool, OmniViz Pro,
Hypercube, play an important role in analyzing clusters. Integrating these
techniques with neural networks can provide interesting results.

GRB classification~\cite{p16} could be an case study to use as a
benchmark. Possible applications could be tested on data sets from the GRB
catalogs, for example using light curves or band-spectral parameters.

Separation of gamma from hadrons is another important and difficult problem in
Gamma-Ray experiments. The classification problem has been addressed with
supervised neural networks. The network separation is based on the study of
simulated data. It is very likely that severe adjustments have to be
made to the simulation to better reflect the data, and the network training has
to be redone with the improved simulation. The disadvantage of this approach is
the output ambiguity and the network should be refined constantly to improve the 
separation of the output. Applying Self-Organizing Networks would be useful as the
classification could be automatic and model-independent.

The final research perspective is a library of Science Tools for AstroParticle
Physics.  Such library should include tools for data mining, tools for
optimizing the features selection (physical characteristics which can
be extracted from different detectors, in particular GLAST, MAGIC, and X-ray
detectors like INTEGRAL, CHANDRA, SWIFT), and a powerful visualization package.


\begin{thebibliography}{99}
  

\bibitem{del92} DELPHI Coll., Phys. Lett. B295 (1992) 383;\\
  L. Lonnblad, C. Petersen and T. Rognvaldsson, Nucl. Phys. B349 (1991) 675;\\
  C. Bortolotto, A. De Angelis and L. Lanceri, Nucl. Instr. and Methods A306
  (1991) 457.
  
\bibitem{praveen} P. Boinee, A. De Angelis, E. Milotti, ``Automatic
  Classification using Self-Organizing Neural Networks in Astrophysical
  Experiments'', in: S. Ciprini, A. De Angelis, P. Lubrano and O. Mansutti
  (eds.): Proc. of "Science with the New Generation of High Energy Gamma-ray
  Experiments" (Perugia, Italy, May 2003). Forum, Udine 2003, p.177,
  arXiv:cs.NE/0307031.
  
\bibitem{ciccia} C. Cecchi, F. Marcucci, G. Tosti,``An application of the
  Independent Component Analysis methodology to gamma ray astrophysical
  imaging'', in: S. Ciprini, A. De Angelis, P. Lubrano and O. Mansutti (eds.):
  Proc. of "Science with the New Generation of High Energy Gamma-ray
  Experiments" (Perugia, Italy, May 2003). Forum, Udine 2003, p. 168,
  astro-ph/0306563.
  
\bibitem{p2} T.~Kohonen, ``Self-Organizing Maps'', Springer, Berlin (1995).
  
\bibitem{p3} B.~Fritzke, ``Growing self-organizing networks - why?'', ESANN,
  Bruges (1996).

\bibitem{p4} A.~K.~Jain, R.~C.~Dubes, ``Algorithms for Clustering Data'',
  Prentice Hall, Englewood Cliffs, New Jersey (1988) 
\bibitem{p6} B.~Fritzke, ``Kohonen feature maps and growing cell structures - a
  performance comparison'', NIPS, Denver (1992).
  
\bibitem{p7} B.~Fritzke, ``Unsupervised clustering with growing cell
  structures'', IJCNN, Seattle (1991).
  
\bibitem{p8} B.~Fritzke, ``Growing self-organizing networks - history, status
  quo, and perspectives'', in ``Kohonen Maps'', Proceedings of WSOM-99, eds. E.
  Oja {\it et al.}, Elsevier (1999).
  
\bibitem{p16} H.~J.~Rajaneimi, P.~Mahonen, ``Classifying GRB using SOM'',
  Astrophys. J. {566} (2001) 202.
  
\bibitem{p1} http://www.batse.msfc.nasa.gov/batse/grb/catalog/current/
  
\bibitem{frailis} M. Frailis, A. De Angelis, V. Roberto, "Data Management and
  Mining in Astrophysical Databases", in: S. Ciprini, A. De Angelis, P. Lubrano
  and O. Mansutti (eds): Proc. of "Science with the New Generation of High
  Energy Gamma-ray Experiments" (Perugia, Italy, May 2003), p. 157
  [arXiv:cs.DB/0307032]
  
\bibitem{p15} SPIRE, http://www.pnl.gov/infoviz/spire/spire.html

\bibitem{glast} http://glast.gsfc.nasa.gov/
  
\bibitem{magic} http://hegra1.mppmu.mpg.de/MAGICWeb/
  
\bibitem{grb} http://www.batse.msfc.nasa.gov/batse/grb/

\end{thebibliography}
\end{document}